\documentclass{article}
\usepackage{arxiv}

\usepackage[utf8]{inputenc} 
\usepackage[T1]{fontenc}    
\usepackage{hyperref}       
\usepackage{url}            
\usepackage{booktabs}
\usepackage{multirow}
\usepackage{amsfonts}       
\usepackage{nicefrac}       
\usepackage{microtype}      
\usepackage{lipsum}		
\usepackage{graphicx}
\usepackage{amsmath}
\usepackage{amssymb}
\usepackage{bbm}
\usepackage{doi}

\usepackage{titletoc}
\usepackage{tocloft} 
\usepackage[sorting=none, url=false, maxbibnames=5]{biblatex}

\renewbibmacro{in:}{}

\newcommand*\samethanks[1][\value{footnote}]{\footnotemark[#1]}

\makeatletter
\renewcommand*{\@fnsymbol}[1]{\ensuremath{\ifcase#1\or \dagger\or \ddagger\or *\or
   \mathsection\or \mathparagraph\or \|\or **\or \dagger\dagger
   \or \ddagger\ddagger \else\@ctrerr\fi}}
\makeatother

\title{Learning to Predict Global Atrial Fibrillation Dynamics from Sparse Measurements}

\author{
    Alexander Jenkins\textsuperscript{1,2}\thanks{Joint first authors}\,, 
    Andrea Cini\textsuperscript{3}\samethanks\,, 
    Joseph Barker\textsuperscript{2}, 
    Alexander Sharp\textsuperscript{4}, 
    Arunashis Sau\textsuperscript{2}
    \\[1ex]
    \textbf{
    Varun Valentine\textsuperscript{2}, 
    Srushti Valasang\textsuperscript{2}, 
    Xinyang Li\textsuperscript{2}, 
    Tom Wong\textsuperscript{2}, 
    Timothy Betts\textsuperscript{5}
    }
    \\[1ex]
    \textbf{
    Danilo Mandic\textsuperscript{1}, 
    Cesare Alippi\textsuperscript{3,6}\thanks{Joint senior authors}\,, 
    Fu Siong Ng\textsuperscript{2}\samethanks
    }
    \\[2ex]
    \textsuperscript{1}Department of Electrical and Electronic Engineering, Imperial College London, United Kingdom\\
    \textsuperscript{2}National Heart and Lung Institute, Imperial College London, United Kingdom\\
    \textsuperscript{3}The Swiss AI Lab IDSIA USI-SUPSI, Universit\`a della Svizzera italiana, Switzerland\\
    \textsuperscript{4}Department of Engineering Science, University of Oxford, United Kingdom\\
    \textsuperscript{5}Oxford University Hospitals NHS Foundation Trust, United Kingdom\\
    \textsuperscript{6}Dipartimento di Elettronica, Informazione e Bioingegneria, Politecnico di Milano, Italy\\[2ex]
    Correspondence to: \href{mailto:f.ng@imperial.ac.uk}{f.ng@imperial.ac.uk}
}

\date{}

\renewcommand{\shorttitle}{Learning to Predict Global Atrial Fibrillation Dynamics from Sparse Measurements}

\hypersetup{
pdftitle={Learning to predict global atrial fibrillation dynamics from sparse measurements},
pdfsubject={q-bio.NC, q-bio.QM},
pdfauthor={David S.~Hippocampus, Elias D.~Striatum},
pdfkeywords={First keyword, Second keyword, More},
}

\usepackage[acronym, nohypertypes={acronym}]{glossaries}

\newacronym{af}{AF}{Atrial Fibrillation}
\newacronym{grnn}{GRNN}{graph recurrent neural network}
\newacronym{gnn}{GNN}{graph neural network}
\newacronym{rnn}{RNN}{recurrent neural network}
\newacronym{mlp}{MLP}{multilayer perceptron}
\newacronym{mpnn}{MPNN}{message passing neural network}
\newacronym{stgnn}{ST-GNN}{spatio-temporal graph neural network}
\newacronym{mf}{MF}{matrix factorisation}
\newacronym{bi}{Bi}{bidirectional}
\newacronym{cdf}{CDF}{cumulative distribution function}
\newacronym{ks}{KS}{Kolmogorov–Smirnov}
\newacronym{mae}{MAE}{mean absolute error}
\newacronym{mre}{MRE}{mean relative error}
\newacronym{mse}{MSE}{mean squared error}
\newacronym{mape}{MAPE}{mean absolute percentage error}
\newacronym{tpr}{TPR}{true positive rate}
\newacronym[longplural={phase singularities},
            plural={PSs},
            firstplural={phase singularities (PSs)}]{ps}{PS}{phase singularity}

\begin{document}
\maketitle

\begin{abstract}
    Catheter ablation of \gls{af} consists of a one-size-fits-all treatment with limited success in persistent \gls{af}. This may be due to our inability to map the dynamics of \gls{af} with the limited resolution and coverage provided by sequential contact mapping catheters, preventing effective patient phenotyping for personalised, targeted ablation. Here we introduce \textsc{FibMap}, a graph recurrent neural network model that reconstructs global \gls{af} dynamics from sparse measurements. Trained and validated on 51 non-contact whole atria recordings, \textsc{FibMap} reconstructs whole atria dynamics from 10\% surface coverage, achieving a 210\% lower mean absolute error and an order of magnitude higher performance in tracking phase singularities compared to baseline methods. Clinical utility of \textsc{FibMap} is demonstrated on real-world contact mapping recordings, achieving reconstruction fidelity comparable to non-contact mapping. \textsc{FibMap}'s state-spaces and patient-specific parameters offer insights for electrophenotyping AF. Integrating \textsc{FibMap} into clinical practice could enable personalised \gls{af} care and improve outcomes.
\end{abstract}


\section{Introduction}
Atrial Fibrillation (AF) is the most common cardiac arrhythmia with a lifetime risk of 1 in 3 \cite{mou2018lifetime}. \Gls{af} is estimated to affect 37.5 million people worldwide - 0.5\% of the global population - with a 60\% projected rise by 2050 \cite{lippi2021global}. The arrhythmia underpins much of global morbidity and mortality as a major cause of stroke \cite{katsanos2020stroke}, heart failure \cite{maisel2003atrial} and death \cite{wolf1998impact}. Global healthcare systems experience a significant and rising cost burden managing AF, with direct patient costs alone estimated to be 2.4\% of the United Kingdom’s £182 billion healthcare budget \cite{2004cost}.

Catheter ablation is a common treatment for \gls{af} \cite{brachmann2021atrial, di2016ablation} and aims to aid the maintenance of sinus (normal) rhythm through the controlled destruction of cardiac tissue via heating or freezing \cite{kuck2016cryoballoon}. In this way, it is possible to electrically isolate regions of the atria that initiate or sustain \gls{af}. Pulmonary vein isolation, which isolates \gls{af} triggers in the pulmonary veins from the left atrium, is central to \gls{af} ablation, yet its success rates are as low as 50\% at 5 years in persistent \gls{af}, with effectiveness diminishing over time \cite{scherr2015five-year}. Other ablation targets, such as linear lesions to compartmentalise the atria and targeted ablation of putative \gls{af} drivers have not significantly improved outcomes \cite{lee2019electrical, verma2015approaches, vogler2015pulmonary, narayan2012treatment, wong2015benefit}. These approaches fail to tailor interventions to the unique electrophenotypes within the spectrum of \gls{af} observed in patients, leaving a one-size-fits-all treatment plan for a highly heterogeneous disorder \cite{ng2020mechanism-directed}. 

\begin{figure}[t]
    \centering
    \includegraphics[width=\linewidth]{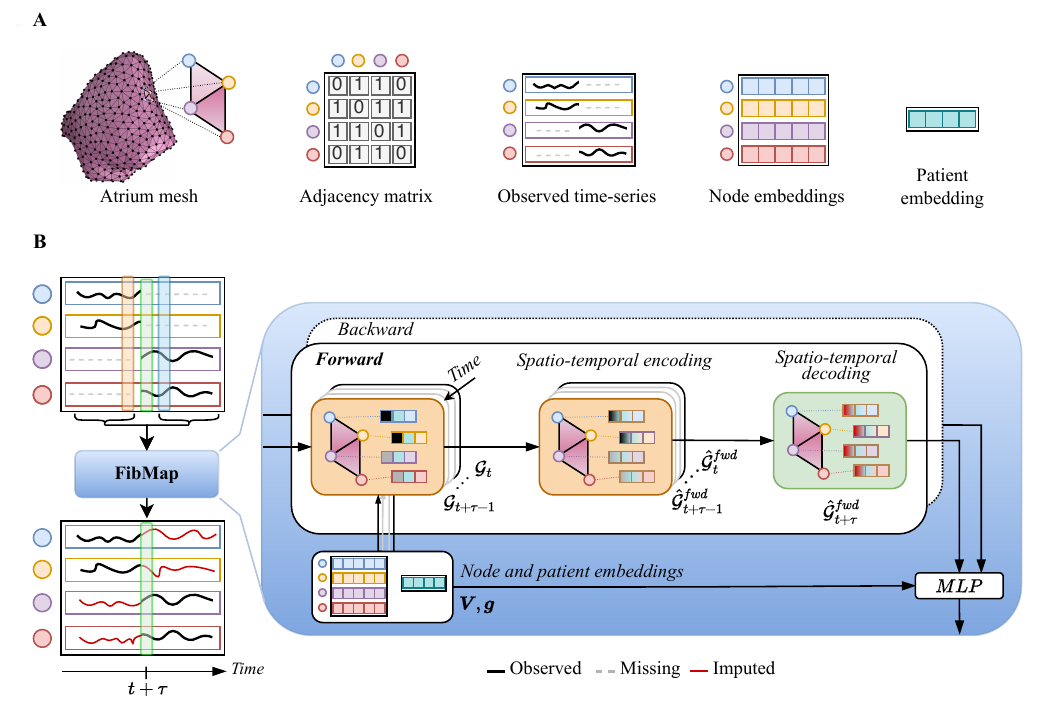}
    \caption{The inputs (A) and architecture (B) of \textsc{FibMap}. A) The atrium is discretised into nodes and edges via a triangulated mesh, and from this the graph adjacency matrix (describing the coupling between nodes) and the observed time series are derived. The patient-specific parameters (node and patient embeddings) are also provided as input to personalise \textsc{FibMap}’s imputation maps. B) \textsc{FibMap} is instantiated as a bidirectional graph recurrent neural network (GRNN), a state-space model that learns a representation for each node by propagating information forwards and backwards through space and time. Node and patient embeddings are provided to a multilayer perceptron (MLP) decoder, which personalises how the learnt representation is mapped to a signal value.}
    \label{fig:fibmap_1}
\end{figure}

The major limitation in applying tailored ablation strategies within the invasive electrophysiology laboratory is our inability to map \gls{af} effectively, that is, reconstruct \gls{af} dynamics on the surface of the heart. Sequential contact mapping is routinely used to record electrical signals from the atria, wherein a multipolar electrode catheter is inserted into the atrial chamber and placed in contact with the endocardial surface during the invasive procedure. The chamber is scanned sequentially with voltage recorded as a function of time and position, and the results are computationally stitched together. This approach is powerful for organised arrhythmias, such as focal or re-entrant atrial tachycardias, where the regularity of the dynamics allows for accurate stitching and thereby precise localisation of the arrhythmic source \cite{ozgul2023high}, resulting in a curative targeted ablation procedure. However, this approach is ineffective in \gls{af} due to its highly disorganised nature, with beat-to-beat variations in wavefront propagation, meaning the non-continuous recordings cannot be sensibly stitched together. While there have been attempts at continuous whole atria mapping, these typically require different modalities such as non-contact catheters that are not routinely used in the clinic since they are expensive, often lack the spatial resolution and coverage necessary to effectively target \gls{af} drivers, and increase procedural complexity and risk \cite{narayan2012treatment, buch2016long-term, rudy1999noninvasive, roney2017spatial}.

The novel approach outlined here is to consider the atria in \gls{af} as a network of oscillators coupled through a graph structure, where non-overlapping regions of tissue are represented as nodes and their coupling to neighbouring tissue via edges \cite{mirollo1990synchronization, aon2006fundamental, qu2014nonlinear}. In doing so, \glspl{gnn} \cite{scarselli2008graph, bronstein2017geometric, bacciu2020gentle}, the generalisation of deep learning methods to irregular (non-Euclidean) data, can be applied to model the dynamics of \gls{af} and reconstruct them from measurements of sequential contact mapping. This method, which we term \textit{imputation mapping}, aims to reconstruct the dynamics of \gls{af} via imputation. By leveraging a \gls{grnn} \cite{seo2018structured, cini2021filling}, a non-linear state-space model suited for systems of coupled oscillators and spatiotemporal data analysis, our model, \textsc{FibMap}, offers a solution for imputation mapping that can 1) express the rich dynamics of fibrillation for each patient, 2) effectively reconstruct the \gls{af} dynamics from the sparse and routinely collected sequential contact measurements, and 3) efficiently generalise to new patients for use in a clinical setting. To this end, imputation mapping is designed to reconstruct whole atria dynamics from routine sequentially collected contact mapping measurements, resulting in an integrative solution for \gls{af} mapping that holds the potential to unlock personalised \gls{af} care, enhance clinical outcomes, and reduce the duration, complexity, and risk of the \gls{af} mapping procedure.

\begin{figure}[t]
    \centering
    \includegraphics[width=\linewidth]{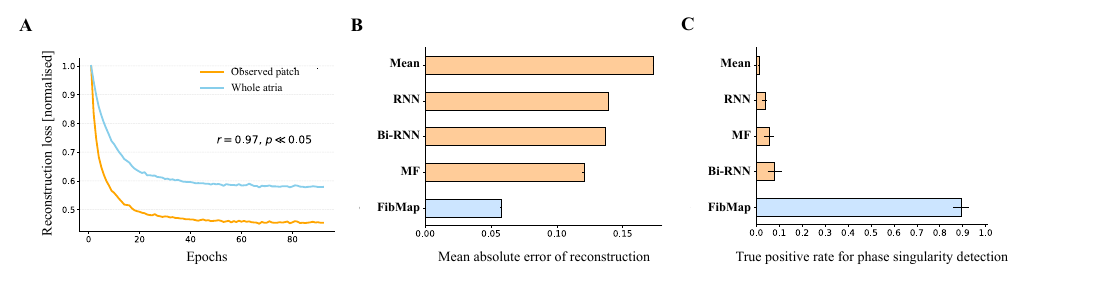}
    \caption{Quantitative results of \textsc{FibMap} imputation mapping. A) Reconstruction loss as a function of the number of epochs for the fine-tuning of \textsc{FibMap} on the validation set, with a strong correlation of $r=0.97$ ($p<0.0001$) present between the loss curves of the observed patch and whole atria. B) Test set reconstruction performance of all models quantified using the mean absolute error (MAE) across all space and time. C) Test set performance for detecting phase singularities (PSs), quantified using the true positive rate of their detection.
}
    \label{fig:fibmap_2}
\end{figure}

\section{Results}
\subsection{\textsc{FibMap} performs accurate whole atria imputation mapping for AF}
Leveraging spatiotemporal message passing \cite{cini2021filling} on a dynamical graph representation of the input, \textsc{FibMap} effectively propagates information from valid measurements across time and the atrial surface to accurately perform imputation mapping. Our solution personalises the reconstruction of \gls{af} dynamics through unique node and patient embedding parameters, akin to word embeddings in natural language processing \cite{li2018word}; conversely, shared parameters are utilised across patients to capture common elements of the dynamics such as the physics of wave propagation. The input representation and architecture of \textsc{FibMap} are detailed in \autoref{fig:fibmap_1}.

Our dataset comprises non-contact recordings from 51 patients with persistent \gls{af}, obtained during an \gls{af} ablation procedure using the AcQMap system before pulmonary vein isolation. The AcQMap system uses non-contact electrodes to sense intra-cavitary unipolar electrograms, from which dipole density measurements are inversely derived to provide improved spatial resolution of electrical activity on the atrial surface \cite{de2022critical, shi2020validation}. Each recording simultaneously samples approximately 3500 nodes across the entire atria at 3000 Hz for 5-20 seconds. Patients were treated between 2016 and 2023 at two centres in the United Kingdom: The Royal Brompton Hospital, London, and the John Radcliffe Hospital, Oxford. The cohort (mean age 64 $\pm$ 11 years, 69\% male) had a mean BMI of 29 $\pm$ 5 kg/m$^2$, with comorbidities including hypertension (39\%), heart failure (33\%), and diabetes mellitus (10\%) (see \autoref{tab:patient_characteristics} in supplementary materials for full details). The procedure involved left atrial mapping, followed by pulmonary vein isolation, and in some cases further adjunctive linear ablations (anterior, posterior, septal) and cardioversion to restore sinus rhythm. These recordings provide a ground truth of human \gls{af} dynamics with fidelity suitable for developing and validating \textsc{FibMap}. After resampling these signals spatially to 500 nodes, temporally to 70 Hz, and normalising the signals to enable consistent analysis of the underlying spatiotemporal patterns independent of measurement units, the dataset was stratified to ensure a spectrum of \gls{af} dynamics within training (70\%), validation (10\%), and test (20\%) sets.

\begin{figure}
    \centering
    \includegraphics[width=\linewidth]{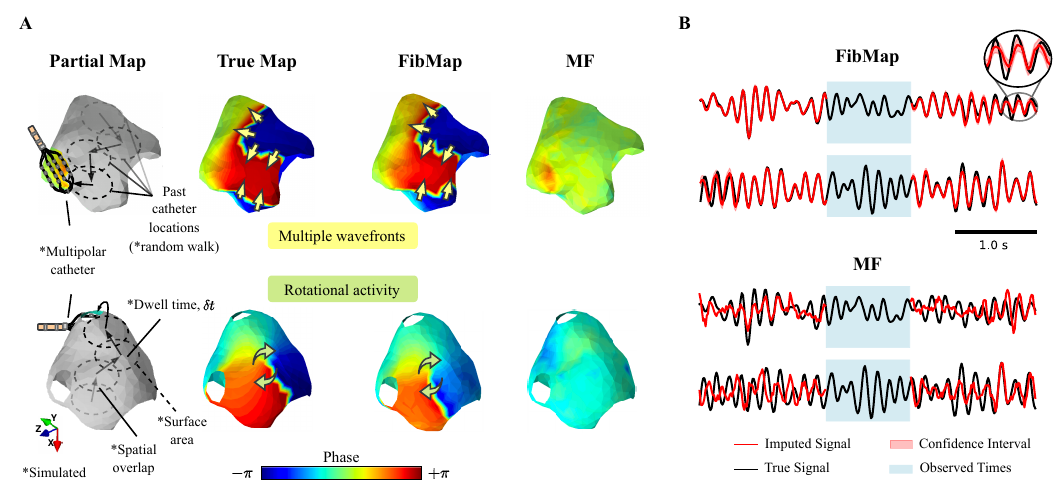}
    \caption{Qualitative results of \textsc{FibMap} imputation mapping. A) Snapshots of the imputation maps of \textsc{FibMap} and MF, versus the ground truth AcQMap phase maps for two different patients (rows). \textsc{FibMap} imputes the missing signals (grey regions) from sparse observations (10\% of atria) of AcQMap recordings with a simulated sequential contact mapping multipolar catheter (surface area of 10\%, no spatial overlap and 1 second dwell time). B) Node-level temporal reconstructions for two different patients (rows) for \textsc{FibMap} and MF. Both the maps and signal traces demonstrate the superior performance of \textsc{FibMap}.}
    \label{fig:fibmap_3}
\end{figure}

Continuous whole atria mapping with systems such as AcQMap are not routinely used in the clinic. Instead, contact mapping catheters are used to precisely measure electrical signals from the atria, with the limitation that they provide sparse measurements across the surface. To emulate routine clinical data collection within the invasive electrophysiology laboratory, a sequential contact mapping strategy was simulated from the whole atria non-contact recordings as a self-avoiding walk of the multipolar catheter (represented as a patch of observations) on the atrial surface; whereby the catheter surface area, dwell time (duration of recording at each location), and spatial overlap, could be controlled (see \autoref{fig:fibmap_3}A). During training, \textsc{FibMap} was optimised for imputation mapping through a whole atria reconstruction loss and a self-supervised learning approach, wherein various multipolar catheter paths were sampled, and parameters augmented (see \autoref{sec:data_and_implem_training} of supplementary materials for more details). As a result, the pre-trained \textsc{FibMap} model, adept at reconstructing a spectrum of \gls{af} dynamics from sparse measurements, remains robust against variations in multipolar mapping.

\begin{table}[t]
    \centering
    \setlength{\tabcolsep}{7pt}  
    \caption{Imputation results for 10\% observed area, dwell time of 1 second, and no spatial overlap.}
    \label{tab:model_comparison}
    \begin{tabular}{@{}l@{\hspace{12pt}}ccccc@{}}  
        \toprule
        \multirow{2}{*}{Model} & \multicolumn{5}{c}{Metrics} \\
        \cmidrule(l{2pt}r{2pt}){2-6}
        & MAE & MSE & MRE & MAPE & PS TPR \\
        \midrule
        Mean & 0.1734{\tiny$\pm$0.0005} & 0.0507{\tiny$\pm$0.0003} & 35.0833{\tiny$\pm$0.1043} & 47.7258{\tiny$\pm$0.3260} & 0.0126{\tiny$\pm$0.0035} \\[2pt]
        Univariate RNN & 0.1393{\tiny$\pm$0.0001} & 0.0298{\tiny$\pm$0.0000} & 28.1724{\tiny$\pm$0.0213} & 44.0633{\tiny$\pm$0.2098} & 0.0369{\tiny$\pm$0.0121} \\[2pt]
        Univariate Bi-RNN & 0.1368{\tiny$\pm$0.0001} & 0.0290{\tiny$\pm$0.0001} & 27.6553{\tiny$\pm$0.0251} & 42.8590{\tiny$\pm$0.1856} & 0.0803{\tiny$\pm$0.0309} \\[2pt]
        MF & 0.1205{\tiny$\pm$0.0012} & 0.0254{\tiny$\pm$0.0005} & 24.3715{\tiny$\pm$0.2434} & 32.5598{\tiny$\pm$0.5609} & 0.0552{\tiny$\pm$0.0207} \\
        \cmidrule[0.2pt]{1-6}
        \textsc{FibMap} & \textbf{0.0574{\tiny$\pm$0.0005}} & \textbf{0.0069{\tiny$\pm$0.0001}} & \textbf{11.5868{\tiny$\pm$0.1098}} & \textbf{16.7715{\tiny$\pm$0.4589}} & \textbf{0.8924{\tiny$\pm$0.0342}} \\
        \bottomrule
    \end{tabular}
\end{table}

This model is efficiently adapted to new patients through a fine-tuning transfer learning procedure, designed to enable accurate whole atria reconstruction from sequential contact mapping data. Our fine-tuning approach preserves essential knowledge for whole atria reconstruction acquired during training by fixing model parameters, while quickly personalising the model by optimising only the patient-specific parameters using an observed patch reconstruction loss (see \autoref{sec:data_and_implem_training} of supplementary materials for more details). The fine-tuning procedure was configured on the validation set, whereby a strong correlation (Pearson’s $r=0.97$, $p<0.0001$) between the observed patch and whole atria reconstruction loss curves was found (see \autoref{fig:fibmap_2}A). This indicates that our fine-tuning procedure enables accurate whole atria reconstruction and generalises effectively to new patients without overfitting to the signals in the observed patches.
The performance of \textsc{FibMap} imputation mapping on new and unseen patients, through our fine-tuning procedure, was quantified on the patients in the test set. Testing involved simulating sequential contact mapping for test set patients with a catheter surface area of 10\% of the atrium, a dwell time of 1 second, and no spatial overlap between successive patches (as shown in \autoref{fig:fibmap_3}A). \textsc{FibMap} was fine-tuned using the configuration found in validation, with the observed patch reconstruction loss monitored to perform early stopping. Test set fine-tuning took a total of 3 hours and 40 minutes, averaging just 22 minutes per patient. After fine-tuning, \textsc{FibMap}’s performance was confirmed through various metrics against ground truth whole atrium signals, all computed on the normalised signals. \textsc{FibMap} significantly outperformed baseline models in predicting whole atrium signals from simulated sequential contact mapping measurements. \autoref{fig:fibmap_2}B illustrates these performances, with the complete set of results shown in Table 1. Specifically, \textsc{FibMap} achieved an \gls{mae} of $0.0574 \pm 0.0005$, a 2.1x improvement compared to the best baseline imputation model, \gls{mf}, which had an \gls{mae} of $0.1205 \pm 0.0012$. In \gls{mse} and \gls{mape}, \textsc{FibMap} scored $0.0069 \pm 0.0001$ and $16.7715 \pm 0.4589$, respectively, compared to \gls{mf}’s $0.0254 \pm 0.0005$ and $32.5598 \pm 0.5609$.

In addition, the performance of the imputation models was assessed by their ability to track the \glspl{ps} present in the recordings, a characteristic of fibrillation dynamics \cite{li2019standardised}. These points, where the phase of electrical activity is undefined, mark the tips of rotating spiral waves that could drive and maintain \gls{af}, representing potential ablation targets and providing crucial insights into the underlying fibrillation mechanisms. Two independent observers manually labelled the \glspl{af} in the central 1-second segment (70 frames) of each map from the different imputation models and patients. \textsc{FibMap} excelled in tracking \glspl{ps}, achieving a \gls{tpr} of $0.8924 \pm 0.0342$, an $11.1\times$ improvement upon the next closest competitor \gls{tpr} of $0.0803 \pm 0.0309$, as illustrated in \autoref{fig:fibmap_2}C and \autoref{tab:model_comparison}. These results demonstrate \textsc{FibMap}’s capability to accurately reconstruct \gls{af} dynamics across space, time, and patients, significantly improving upon existing imputation methods.

Our empirical results confirm \textsc{FibMap}’s capability to reconstruct the complex dynamics of \gls{af} accurately. \autoref{fig:fibmap_3}B illustrates \textsc{FibMap}’s signal reconstructions at individual nodes compared to baseline methods (columns) across two test patients (rows), highlighting its ability to reconstruct the temporal dynamics of \gls{af}. While small amplitude variations are observed between \textsc{FibMap}’s reconstruction and the ground truth recordings, the true signal often resides within \textsc{FibMap}’s predicted confidence intervals (90th – 10th quantiles, see \autoref{sec:fibmap_optimisation} of supplementary materials for more details). Furthermore, \textsc{FibMap}’s reconstruction maintains phase coherence with the ground truth recordings, allowing for the accurate visualisation of phase maps across the atrium.

\autoref{fig:fibmap_3}A provides a snapshot of \textsc{FibMap}’s phase maps derived from the predicted signal, demonstrating its precise reconstruction of \gls{af} spatial dynamics for two different patients (rows). These maps reveal characteristic features of complex \gls{af} dynamics such as multiple wavefronts and rotational activity that were otherwise not visible in the observed patches. In comparison, our empirical findings in \autoref{fig:fibmap_3}B underscore the limitations of the baseline models. While \gls{mf} could in principle be used here, it fails in capturing the complex spatiotemporal dynamics that characterise \gls{af}, as evidenced by reconstruction errors in both amplitude and phase in \autoref{fig:fibmap_3}B which result in poor phase map reconstruction in \autoref{fig:fibmap_3}A.

\subsection{\textsc{FibMap} reconstructs personalised AF dynamics from clinical sequential contact mapping data}
\textsc{FibMap}'s ability to reconstruct whole atria dynamics from real-world sequential contact mapping data was validated retrospectively using recordings from the Advisor\textsuperscript{\texttrademark} HD Grid Mapping catheter (16 electrodes arranged in a grid and evenly spaced 3mm apart, referred to as HD Grid) and the EnSite Precision mapping system (Abbott Laboratories, Lake County, Illinois, United States). This system is routinely used for mapping cardiac arrhythmia and guiding ablation procedures in the invasive electrophysiology laboratory. For three \gls{af} patients in the test set, both EnSite Precision contact mapping and non-contact AcQMap recordings of different durations (tens of minutes vs. seconds) were collected in sequence (non-contemporaneously) before ablation was performed. EnSite Precision HD Grid recordings were resampled to 500 nodes at 70 Hz, electrograms were normalised, and imputation maps were created from the EnSite Precision HD Grid using our fine-tuning procedure (see \autoref{sec:data_and_implem_hd_grid} of supplementary materials for more details).

\begin{figure}[t]
    \centering
    \includegraphics[width=\linewidth]{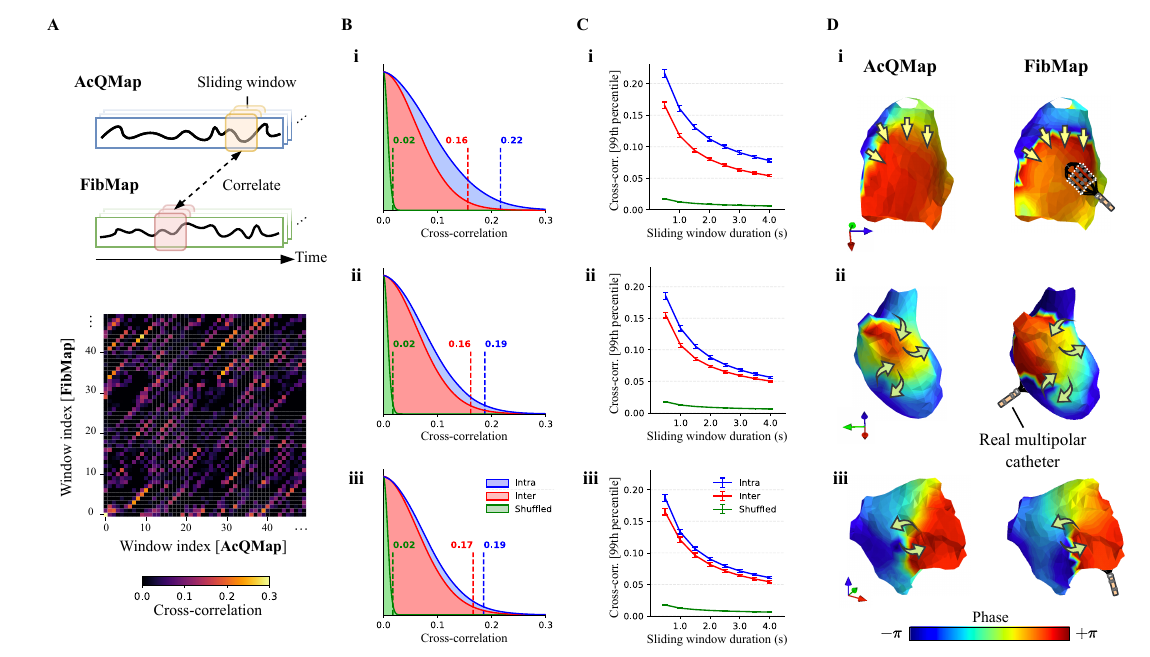}
    \caption{Validation of \textsc{FibMap} imputation maps from EnSite Precision HD Grid Mapping against non-contemporaneous ground truth AcQMap recordings. A) Sliding window cross-correlation analysis between AcQMap and \textsc{FibMap} phase signals, enabling comparison between non-contemporaneous recordings by measuring the similarity between AF patterns across different time windows. The correlation matrix below shows pairwise comparisons between all possible window combinations. For three patients (indexed i-iii), B) shows the kernel density estimates of pairwise cross-correlations computed between AcQMap and \textsc{FibMap} maps from the same patient (intra), different patients (inter), and spatiotemporally shuffled maps (random baseline), using 0.5-second sliding windows. Vertical dashed lines indicate the 99th percentile, with consistently higher values for intra-patient (0.19-0.22) versus inter-patient (0.16-0.17) correlations and shuffled baseline (0.02), demonstrating patient-specific pattern capture. C) Temporal robustness analysis showing the 99th percentile of cross-correlations against sliding window duration. Non-overlapping confidence intervals (computed via bootstrap sampling) between intra-patient and other distributions confirm that \textsc{FibMap} can capture patient-specific dynamics across different temporal scales. D) Representative snapshots of phase maps from highly correlated windows, comparing AcQMap recordings with \textsc{FibMap} maps derived from real HD Grid catheter measurements covering $<10$\% of the atrial surface per time. The maps demonstrate the ability of \textsc{FibMap} to capture AF dynamics which are not directly visible in the original HD Grid measurements, including multiple wavefronts and rotational patterns.
}
    \label{fig:fibmap_4}
\end{figure}

A comparative approach was developed to assess \textsc{FibMap} imputation maps from the EnSite Precision HD Grid against the whole atria `ground truth' provided by AcQMap. The non-contemporaneous nature of these recordings and their varying durations present analytical challenges. \gls{af} wavefront patterns vary beat-to-beat, preventing direct temporal alignment and direct comparison between the imputed maps from the EnSite Precision HD Grid versus the ground truth of the global AcQMap recordings (which were collected minutes apart). To overcome this, a sliding window analysis was implemented (\autoref{fig:fibmap_4}A) to cross-correlate segments of ground truth AcQMap and imputed \textsc{FibMap} phase signals, quantifying the overall dynamical similarity between these non-contemporaneous recordings, without requiring exact temporal matching. This approach enables fair and meaningful comparison between the two data modalities (imputed maps vs. ground truth). The resulting cross-correlation distributions were analysed to determine whether \textsc{FibMap} captures patient-specific dynamics and distinguishes genuine electrophysiological patterns from arbitrary correlations expected by random chance.

To investigate patient specificity, distributions of cross-correlation were computed by comparing AcQMap recordings (ground truth) to \textsc{FibMap} (imputed maps) of the same (intra-) patient and different (inter-) patients using a sliding window of duration 0.5 seconds. To compute the cross-correlation, recordings were projected between source and target surfaces using a nearest-neighbours approach (see \autoref{sec:data_and_implem_finetuning} of supplementary materials for more details). Cross-correlation distributions were plotted as kernel density estimates in \autoref{fig:fibmap_4}Bi-iii for all three patients, revealing consistently wider tails for intra-patient correlations between AcQMap and \textsc{FibMap}, with 99th percentiles ranging from 0.19-0.22 compared to 0.16-0.17 for inter-patient correlations. The higher frequency of strong correlations within the same patient demonstrates \textsc{FibMap}'s ability to capture patient-specific dynamics rather than generic patterns common across patients.

To assess whether these correlations were meaningful, we compared them against a random baseline created by spatiotemporally shuffling the \textsc{FibMap} imputation maps and then performing an intra-patient comparison with AcQMap. The significant separation between intra-patient and random distributions (99th percentiles shown by vertical dashed lines in \autoref{fig:fibmap_4}Bi-iii) confirms that the similarities between AcQMap and \textsc{FibMap} are not due to chance, validating \textsc{FibMap}’s ability to impute \gls{af} dynamics. If this were due to chance, the observed 99th percentile for intra-patient correlations would lie at approximately 0.02 for each patient, an order of magnitude lower than the observed values of 0.19-0.22 (see \autoref{fig:fibmap_4}Bi-iii).

To test the robustness of these findings, sliding window durations were varied from 0.5 to 4.0 seconds (\autoref{fig:fibmap_4}Ci-iii). The 99th percentiles of the cross-correlation distributions were plotted against window duration, showing that while these values generally decreased with longer windows, the 99th percentiles for intra-patient correlations consistently remained higher than both inter-patient and shuffled comparisons. For each patient, confidence intervals were computed via bootstrap sampling (see \autoref{sec:data_and_implem_hd_grid} of supplementary materials for more details), revealing non-overlapping intervals between intra-patient, inter-patient, and shuffled distributions across all window durations, confirming the statistical significance of these differences. Representative phase maps (\autoref{fig:fibmap_4}D) visually confirm the agreement between \textsc{FibMap} imputations and AcQMap ground truth recordings to reveal complex \gls{af} dynamics such as multiple wavefronts and rotational activity, which were otherwise not visible by the EnSite Precision HD Grid which covers less than 10\% of the atrial surface.

These findings demonstrate that \textsc{FibMap}, through our fine-tuning procedure, can reconstruct personalised whole atria \gls{af} dynamics from real examples of multipolar catheter contact recordings collected in the invasive electrophysiology laboratory with fidelity comparable to AcQMap. This demonstrates the capacity of imputation mapping to integrate with and significantly enhance the capabilities of routinely used multipolar catheter mapping systems.

\subsection{The sensitivity of \textsc{FibMap} to variations in multipolar mapping and AF organisation}
Returning to our analysis of simulated sequential contact mapping, a sensitivity analysis was conducted to assess \textsc{FibMap}’s practical application by evaluating its reconstruction performance of whole atria dynamics across various catheter areas, dwell times, and levels of \gls{af} organisation. Spatial overlap was fixed at zero for this analysis. \gls{af} organisation was quantified by computing the average Shannon entropy across the ground truth signals of each patient, where lower values indicate more organised dynamics. Patients were then divided into two groups of low and high entropy based on the median average Shannon entropy value.

\begin{figure}
    \centering
    \includegraphics[width=\linewidth]{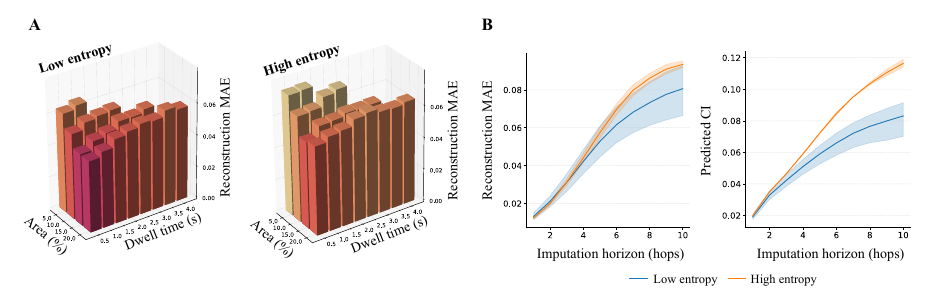}
    \caption{Results of \textsc{FibMap} sensitivity analysis. A) MAE of \textsc{FibMap} reconstructions as a function of catheter surface area, dwell time and entropy (left vs. right). B) Reconstruction MAE of \textsc{FibMap} and the predicted confidence intervals as a function of imputation horizon.}
    \label{fig:fibmap_5}
\end{figure}

\autoref{fig:fibmap_5}A illustrates the \gls{mae} of the whole atria reconstruction for both groups across varying catheter areas and dwell times. The low entropy group (more organised forms of \gls{af}) consistently exhibited lower \gls{mae} values, with 0.065 on average compared to 0.071 of the high entropy group. Furthermore, both groups demonstrated improved reconstruction performance with increased catheter surface area and reduced dwell time, achieving a minimal \gls{mae} of 0.045 in the low entropy group with a 20\% catheter surface area and a 0.5s dwell time. A shorter dwell time enables faster sampling of the whole atrium, allowing it to be covered multiple times within the finite duration of our recordings. For each combination of dwell time and catheter surface area depicted in \autoref{fig:fibmap_5}A, we ensure that each patient’s atrium is covered at least once. Combinations that do not meet this criterion are not shown. While these results could guide the optimal use of \textsc{FibMap}, we were unable to evaluate its performance for longer durations due to the limited length of our recordings. However, in practice, sampling duration is not restricted.

Performance was assessed based on the imputation horizon to reveal insights into \textsc{FibMap}’s capabilities and their implications for practical application. The imputation horizon, defined as the spatiotemporal distance from the closest observed node and time point (measured in hops on a spatiotemporal graph), measures how far (in space and time) \textsc{FibMap} can reliably predict beyond the observed data points. \autoref{fig:fibmap_5}B demonstrates that the high entropy group experienced a sharper increase in \gls{mae} with increasing imputation horizon compared to the low entropy group, where \gls{mae} eventually plateaued, possibly due to \textsc{FibMap}’s predictions losing phase coherence with the ground truth. Concurrently, \autoref{fig:fibmap_5}C demonstrates a corresponding increase in predicted confidence intervals with the imputation horizon. This underscores \textsc{FibMap}’s ability to quantify uncertainty in long-range predictions, which is crucial for reliably interpreting whole atria imputation maps and making informed clinical assessments.

\subsection{Revealing structure in \textsc{FibMap}’s state- and embedding-spaces}
\begin{figure}
    \centering
    \includegraphics[width=\linewidth]{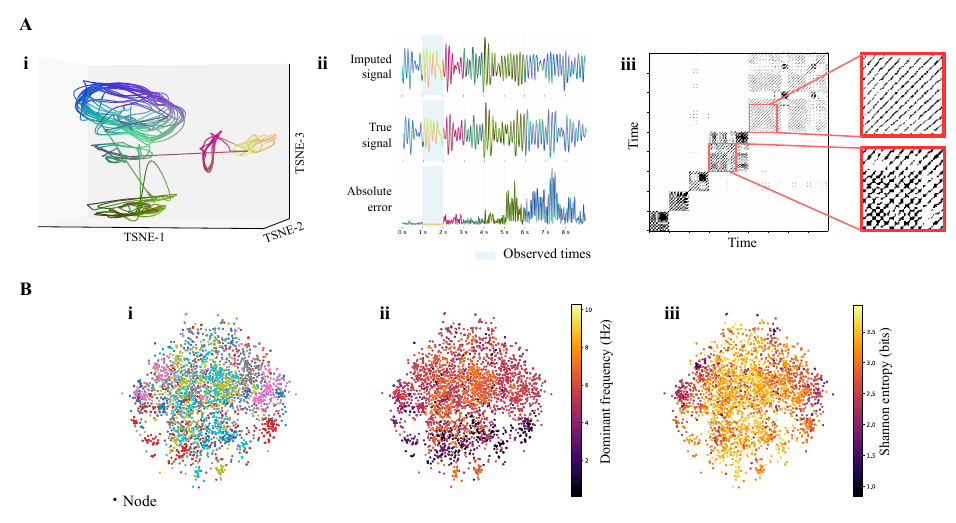}
    \caption{Interpretation of \textsc{FibMap} hidden state and patient-specific node parameters. Ai) Trajectory of the hidden state of a node in the test set, where the state space has been reduced to 3D using t-SNE dimensionality reduction. Trajectories are coloured by their position in the state space by converting the 3D coordinate to RGB. Aii) The imputed and true signals are plotted for the same node, coloured by their position in the state space. The absolute error between the imputed and true signals is also shown. Aiii) A recurrence plot summarising the state-space trajectory, where different dynamical structures are observed over time. B) Node embeddings in the test set coloured by (i) patient, (ii) dominant frequency and (iii) Shannon entropy.}
    \label{fig:fibmap_6}
\end{figure}

\textsc{FibMap}'s design as a non-linear state-space model enables the visualisation of hidden state dynamics across all nodes and times. These hidden states exhibit chaotic dynamics, characterised by diverging trajectories around fixed points in state space, effectively capturing the inherent chaos of \gls{af}. In \autoref{fig:fibmap_6}Ai, we use t-SNE dimensionality reduction \cite{van2008visualizing} to visualise the state-space trajectory of a given node in a 3D space. Converting these 3D coordinates to RGB values creates a colour-coding that reveals orbital patterns and transitions across different regions of state space.

To examine the relationship between state space regions and signal dynamics, \autoref{fig:fibmap_6}Aii shows the imputed and ground truth signals coloured by their state space positions, along with their absolute error to reveal state-dependent variations. While both state duration and absolute error vary, the dynamics show subtle changes between states. Figure 6Aiii presents a recurrence plot \cite{marwan2007recurrence} of the state-space trajectory, revealing structured dynamics within and between states. For example, parallel lines to the main diagonal appear at later times, which is a characteristic of deterministic dynamics where trajectories repeatedly progress through the same sequence of states.

While a comprehensive analysis of state space dynamics exceeds the scope of this work, our findings demonstrate that \textsc{FibMap}’s formulation as a state-space model inherently captures important information about \gls{af} dynamics. Moreover, non-linear dynamics analysis tools, such as recurrence plots, can extract additional insights from \textsc{FibMap}. These insights could enable electrophenotyping of \gls{af} patients into broad subgroups, help predict the probability of benefit from ablation, and guide ablation strategy.

The learned patient and node embedding parameters in \textsc{FibMap} provide additional structural insights. \autoref{fig:fibmap_6}B visualises the node embeddings for each patient in the test set in 2D space using t-SNE dimensionality reduction \cite{van2008visualizing}. \autoref{fig:fibmap_6}Bi shows each node (marker) coloured uniquely for each patient, revealing both localised clusters and nodes distributed across the space, suggesting similarities (and differences) between the nodes of different patients. To interpret these embeddings, \autoref{fig:fibmap_6}Bii and \autoref{fig:fibmap_6}Biii show the dominant frequency and Shannon entropy of the signal at each node, respectively. The node embeddings capture this information in a complex manner, with dominant frequency generally decreasing toward the bottom and Shannon entropy increasing toward the centre. These node embedding spaces could enable a systematic comparison of tissue properties across patients and anatomical locations, while patient embeddings could identify individuals with similar global dynamics.

These findings indicate that both \textsc{FibMap}’s hidden states and embedding parameters could aid in electrophenotyping patients. However, since follow-up information or spatial properties of the tissue were not available for our dataset, further research is needed to fully interpret and validate the information contained within the patient and node embeddings.




\section{Discussion}
The inability to map \gls{af} effectively in the invasive electrophysiology laboratory limits the application of tailored ablation strategies and potentially hinders curative treatment outcomes. To address this problem, we introduce imputation mapping to reconstruct full \gls{af} dynamics from recordings collected through non-continuous sequential contact mapping, which observe only a fraction of the atrium at a time. Until now, these sparse measurements could not be effectively stitched into a global map due to the highly disorganised nature of \gls{af}. To this end, our solution, termed \textsc{FibMap}, employs a novel \gls{grnn}-based non-linear state-space model for personalised reconstruction of global \gls{af} dynamics.

\textsc{FibMap} was trained and rigorously validated on a dataset of 51 persistent \gls{af} patients, using continuous non-contact whole atria recordings from the AcQMap system as ground truth and simulated sequential contact mapping to replicate clinical data collection. Our results show that \textsc{FibMap} outperforms the baseline imputation models in whole atria reconstruction, achieving over $2\times$ improvement in \gls{mae} and more than an order of magnitude improvement in \gls{ps} detection rates, while observing only 10\% of the atrium. Empirical results demonstrate that \textsc{FibMap} accurately captures both amplitude and phase relations across the atria, reconstructing complex features of \gls{af} dynamics such as rotational activity and multiple wavefronts. Beyond simulated contact mapping, we showed that \textsc{FibMap} successfully transfers to real clinical examples of sequential contact mapping, collected with the multipolar HD Grid catheter, to reconstruct \gls{af} dynamics with fidelity comparable to global AcQMap. This demonstrated the ability of \textsc{FibMap} to integrate sparse contact recordings of \gls{af} into a coherent imputation map. We expect further improvements in reconstruction quality when the catheter placement is guided prospectively (possibly by the results of our sensitivity analysis). Finally, visualisation of \textsc{FibMap}’s hidden state and embedding spaces reveals significant structure, suggesting potential for future electrophenotyping studies. 

In summary, \textsc{FibMap} provides a comprehensive solution for imputation mapping by: 1) capturing the rich dynamics of fibrillation for each patient, 2) effectively reconstructing whole atria maps from sequential contact measurements, 3) producing uncertainty estimates to facilitate a clinical interpretation, 4) efficiently generalising to new patients for clinical use through our fine-tuning procedure, and 5) parameterising \gls{af} phenotypes, offering the opportunity for downstream work to investigate its potential in personalising \gls{af} care.

While methods for whole atria mapping do exist, such as the AcQMap system and non-invasive electrocardiographic imaging, these approaches have not demonstrated sufficient efficacy for routine clinical use, due in part to their non-contact nature \cite{narayan2012treatment, buch2016long-term, rudy1999noninvasive, roney2017spatial}. These non-contact methods suffer from having to calculate the electrograms on the heart surface rather than recording true contact electrograms. As a result, global non-contact mapping approaches suffer from low spatial resolution, leading to incorrect interpretation of \gls{af} dynamics \cite{roney2017spatial}. Imputation mapping offers a novel approach to obtaining whole atria maps from routinely collected sequential contact mapping data, efficiently integrating with existing mapping multipolar catheters (such as EnSite Precision HD Grid) to reconstruct \gls{af} dynamics in a real-time or post hoc manner. Although our work has demonstrated a proof-of-concept for imputation mapping, future clinical studies should focus on validating its efficacy in guiding personalised ablative strategies. These studies should also determine the optimal spatial and temporal resolution of the imputation map for accurate identification of fibrillation drivers and their precise ablation.

A crucial design choice in our study was the use of data from the AcQMap system. Non-contact data provide a realistic ground truth for human \gls{af}, which is essential for accurate model training and validation. Using these measurements, \textsc{FibMap} can learn an empirical model of \gls{af} dynamics, ensuring an accurate representation of human \gls{af}. In contrast, using simulated whole atria signals would bias the model towards simplified and debated mathematical models of fibrillation \cite{fenton1998vortex, clayton2011models}. While this approach proved effective for our proof-of-concept validation on AcQMap and EnSite Precision HD Grid recordings, if future work did identify limitations, \textsc{FibMap} can be trained on a mixture of real and simulated fibrillation data to enhance its robustness. However, we emphasise that the current design of \textsc{FibMap} incorporates a pre-trained model component that permits continual learning, improving accuracy over time as it is applied to and fine-tuned on contact mapping recordings from new patients across the spectrum of \gls{af}. Imputation mapping, as demonstrated by \textsc{FibMap}, offers a promising approach for reconstructing whole atria dynamics from routinely collected sparse sequential contact mapping data.

Future research will focus on proving the efficacy of imputation mapping in guiding personalised catheter ablation procedures and determining the optimal resolution requirements for its successful clinical deployment. Additionally, refinements to the architecture of \textsc{FibMap} will further improve imputation accuracy. For example, hierarchical architectures \cite{cini2023graph-based, marisca2022learning} would improve the imputation performance at longer horizons. By effectively bridging the gap between sequential contact mapping and continuous whole atria mapping systems, imputation mapping allows clinicians to \emph{see more with less}. Its integration with existing contact mapping systems has the potential to transform our interpretation of complex \gls{af} dynamics, ultimately improving the precision and effectiveness of \gls{af} ablation.

\section*{Acknowledgments}
AJ was supported by the UKRI CDT in AI for Healthcare \url{http://ai4health.io} (Grant No. P/S023283/1). JB was supported by the British Heart Foundation (FS/CRTF/24/24624). AS received a research grant and travel support from Acutus Medical. TB received honoraria from Acutus Medical. CA was partly funded by the Swiss National Science Foundation under grant 204061: High-Order Relations and Dynamics in Graph Neural Networks. FSN was supported by the British Heart Foundation (RG/F/22/110078, RE/19/4/130023) and the National Institute for Health Research Imperial Biomedical Research Centre.

\section*{Author contributions}
AJ conceptualised the solution, designed and performed the experiments, and wrote the manuscript. AC conceptualised the solution, designed the experiments, and edited the manuscript. JB designed the experiments and wrote and edited the manuscript. VV and SV manually annotated the phase singularities present in the recordings. A Sharp, A Sau, XL, TW, and TB supplied clinical data for analysis. DM, CA and FSN co-supervised the design of the solution and the empirical evaluation and edited the manuscript.

\section*{Competing interest declaration}
AJ, AC, DM, CA and FSN are applicants on the patent: \textit{A Method of Constructing Maps of Dynamical Variables on a Cardiac Surface} (UK Patent Application No. 2500830.1). All other authors declare no conflicts of interest.

\AtNextBibliography{\small}
\printbibliography[heading=bibintoc]
\newpage


\section*{Supplementary materials}

\setcounter{section}{0}
\renewcommand{\thesection}{S.\arabic{section}}

\renewcommand{\thesubsection}{\thesection.\arabic{subsection}}
\renewcommand{\thesubsubsection}{\thesubsection.\arabic{subsubsection}}

\startcontents[appendices]
\printcontents[appendices]{l}{1}{}

\setcounter{figure}{0}
\setcounter{table}{0}
\renewcommand{\thefigure}{S.F\arabic{figure}}
\renewcommand{\thetable}{S.T\arabic{table}}

\section{Background and related work}
\label{sec:background}
The field of graph signal processing \cite{8347162, mandicFTMLp2} has emerged in the last decade to generalise digital signal processing methods, such as convolutional filters, to the graph domain. A graph, $\mathcal{G}(\mathcal{V}, \mathcal{E})$, is defined as a set of $N$ nodes, $v_i \in \mathcal{V}$ for $i = 1, \ldots, N$, and a set of $\mathcal{E}$ edges denoting the pairwise connections between them, $e_{ij} = (v_i, v_j) \in \mathcal{E}$, for $i = 1, \ldots, N$ and $j = 1, \ldots, N$. Graph signal processing focuses on analysing signals on graphs, where a multivariate signal on a graph is defined as $\mathbf{X} \in \mathbb{R}^{N\times c}$, which assigns a real-valued signal of $c$ channels to each node, $\mathbf{X}: \mathcal{V} \mapsto \mathbb{R}^c$.

Building on graph signal processing methods, graph deep learning \cite{bronstein2017geometric, mandicFTMLp3} has generalised successful deep learning architectures, such as convolutional neural networks, to the graph domain \cite{kipf2016semi-supervised, zhang2020deep}. \Glspl{stgnn} refer to the class of deep learning architectures designed to analyse time-varying graph signals \cite{cini2023graphdeeplearningtime, jin2024survey}. \Glspl{stgnn} can be categorised into time-then-space \cite{gao2022equivalence, cini2023scalable} or time-and-space models \cite{seo2018structured, li2018diffusionconvolutionalrecurrentneural, marisca2022learning}, which denote separate or joint processing of the space and time dimensions, respectively. A notable example of a time-and-space \gls{stgnn} is the \gls{grnn}, which replaces the fully connected layers in a \gls{rnn} with graph convolutions \cite{seo2018structured, li2018diffusionconvolutionalrecurrentneural}. \Glspl{stgnn} have achieved state-of-the-art performance in tasks such as forecasting \cite{jiang2022graph, lam2023learning, cini2023graph-based} and missing data imputation \cite{marisca2022learning, cini2021filling}.

\glspl{stgnn} are inherently global models, sharing parameters across space and assuming a stationary process over time. Global \glspl{stgnn} can be used for zero-shot transfer and inductive learning on unseen graphs, however, they might fail to account for spatial variations in dynamics. Node embeddings have been recently introduced to learn these local effects in ST-GNNs \cite{cini2024taming}. Whilst hybrid global-local models often outperform global architectures, recent research has focussed on improving their performance in an inductive setting using transfer learning \cite{cini2024taming, yin2022nodetrans}.

Other imputation methods have been applied to time series, ranging from \gls{mf} methods \cite{salakhutdinov2008bayesian, lee2000algorithms} and their counterparts with graph and temporal regularisation \cite{cai2010graph, yu2016temporal}, to deep learning techniques employing \glspl{rnn} \cite{cao2018brits}, generative adversarial networks \cite{yoon2019time, liu2019naomi}, transformers \cite{du2023saits}, or most recently denoising diffusion probabilistic models \cite{tashiro2021csdi, alcaraz2021diffusion, jenkins2023improving}.

\section{\textsc{FibMap}}
\label{sec:fibmap}
\textsc{FibMap} aims to reconstruct the dynamics of \gls{af} from partial measurements collected during sequential contact mapping using a \gls{grnn}. This task is formulated using a spatiotemporal time series imputation framework, which models the propagation of electrical signals across the atrium via spatiotemporal message passing on a graph.

To become a feasible clinical solution, the design of \textsc{FibMap} solves several technical challenges of the task: 1) how to effectively propagate information from highly sparse sets of observations where up to 10\% of the atrium is observed per time with sequential contact mapping; 2) how to manage possibly unknown parameters which could affect the dynamics, such as the spatial properties of the tissue and patient-specific covariates; 3) how to efficiently train and use a single model within the duration of the clinical procedure, whilst balancing the trade-off between personalised reconstruction and generalising across a spectrum of \gls{af} dynamics; and 4) how to quantify the uncertainty of the estimated reconstruction — an essential factor for ensuring a utile and trustworthy clinical tool in practice. In this section, we describe the framework behind \textsc{FibMap}, our solution for achieving accurate reconstruction of \gls{af} dynamics while considering the technical challenges detailed above.

\subsection{Inputs}
\label{sec:fibmap_inputs}
A graph adjacency matrix for each patient $p$, $\mathbf{A}^{(p)} \in \mathbb{R}^{N \times N}$, is created by discretising the atrial surface into $N$ nodes and triangulating the surface to create edges. The adjacency matrix, $\mathbf{A}^{(p)}$, is constant over time. A sequence of graphs is formed as a tuple, $\mathcal{G}_{t:T}^{(p)} = \langle \mathbf{X}_{t:T}^{(p)}, \mathbf{M}_{t:T}^{(p)}, \mathbf{A}^{(p)} \rangle$, whereby partial electrical signals, $\mathbf{X}_{t:T}^{(p)} \in \mathbb{R}^{N \times 1}$, are observed via sequential contact mapping, and the observation mask, $\mathbf{M}_{t:T}^{(p)} \in \{0,1\}^{N\times 1}$, specifies the indices of the valid measurements. 

To manage the patient’s latent spatial and global parameters which could affect the dynamics of \gls{af}, trainable embeddings are used as an additional input. Specifically, trainable node embeddings, $\mathbf{V}^{(p)} \in \mathbb{R}^{N \times q}$, are used to learn the latent spatial properties of the patient’s tissue, and trainable patient embeddings, $\mathbf{g}^{(p)} \in  \mathbb{R}^{r}$, will be used to learn the latent global properties of the patient. In this way, the sequence of graph data for each patient can be expressed as a tuple, $\mathcal{G}_{t:L}^{(p)} = \langle \mathbf{X}_{t:L}^{(p)}, \mathbf{M}_{t:L}^{(p)}, \mathbf{A}^{(p)}, \mathbf{V}^{(p)}, \mathbf{g}^{(p)} \rangle$. For efficient processing, input sequences are formed by splitting the sequence of length, $L$, into windows of length, $W$, processed with a stride, $M$.

\subsection{Architecture}
\label{sec:fibmap_architecture}
\textsc{FibMap} is instantiated as a bidirectional gated-\gls{grnn}, a non-linear state-space model designed to propagate information from valid observations across space and time. Our solution is composed of two spatiotemporal encoder blocks and a decoder. The spatiotemporal encoders operate in two different directions, processing the sequence in both a forward (\textit{fwd}) and backward (\textit{bwd}) direction, respectively. \textsc{FibMap}'s architecture extends the framework proposed in \cite{cini2021filling} by introducing local and global embedding parameters to address the \gls{af} mapping problem. These modifications enable generalisation across patients and efficient transfer to new patients via fine-tuning.

To balance the trade-off between personalised reconstruction and generalisation across a spectrum of \gls{af} dynamics with a single model, \textsc{FibMap} is designed to perform personalised reconstruction by providing patient-specific parameters, $\mathbf{V}^{(p)}$ and $\mathbf{g}^{(p)}$, as additional input to the encoder and decoder components of the architecture. All other model parameters are shared across patients to learn the common aspects of the dynamics (such as the physics of the problem). We begin by defining the spatiotemporal encoder block of the architecture, before explaining the decoder.

\subsubsection{The spatiotemporal encoder}
\label{sec:fibmap_encoder}
First, the observed data at time $t$ is encoded as
\begin{equation}
\mathbf{H}_t^{0} = \text{MLP}_{\text{enc}}([ \mathbf{X}_t^{(p)} \odot \mathbf{M}_t^{(p)} \| \mathbf{M}_t^{(p)} \| \mathbf{G}^{(p)} \| \mathbf{V}^{(p)}]),
\end{equation}
where the symbols $\odot$ and $\|$ denote the Hadamard product and concatenation operator, respectively, and $\mathbf{G}^{(p)} \in \mathbb{R}^{N\times r}$ is a matrix where copies of $\mathbf{g}^{(p)}$ are stacked across nodes. The encoding step constructs a representation of the observed and missing values alongside the patient-specific parameters in a common embedding space, $\mathbf{H}_t^{0} \in \mathbb{R}^{N\times d}$. Next, the embedded data is processed sequentially using a gated-\gls{grnn}, where the $k$-th layer is given by
\begin{subequations}
\begin{align}
    \mathbf{Z}_t^k &= \mathbf{H}_{t-1}^k, \\
    \mathbf{R}_t^k &= \sigma(\text{MP}_r^k([\mathbf{Z}_t^k \| \mathbf{H}_{t-1}^k], \mathcal{E})), \\
    \mathbf{U}_t^k &= \sigma(\text{MP}_u^k([\mathbf{Z}_t^k \| \mathbf{H}_{t-1}^k], \mathcal{E})), \\
    \mathbf{C}_t^k &= \tanh(\text{MP}_c^k([\mathbf{Z}_t^k \| \mathbf{R}_t^k \odot \mathbf{H}_{t-1}^k], \mathcal{E})), \\
    \mathbf{H}_t^k &= \mathbf{U}_t^k \odot \mathbf{H}_{t-1}^k + (1-\mathbf{U}_t^k) \odot \mathbf{C}_t^k.
\end{align}
\end{subequations}
$\text{MP}_r^k(\cdot)$, $\text{MP}_u^k(\cdot)$ and $\text{MP}_c^k(\cdot)$ denote the \gls{mpnn} layers for the reset, update, and candidate gates, respectively, and the activation functions $\sigma(\cdot)$ and $\text{tanh}(\cdot)$ denote the sigmoid and hyperbolic tangents. The hidden state of the gated-\gls{grnn} at each layer $k$ is initialised as a linear function of the node embeddings
\begin{equation}
    \mathbf{H}_{t=0}^k = \mathbf{V}^{(p)} \mathbf{W}_{\text{init}}^k + \mathbf{b}_{\text{init}}^k,
\end{equation}
where $\mathbf{W}_{\text{init}}^k \in \mathbb{R}^{q\times d}$ and $\mathbf{b}_{\text{init}}^k \in \mathbb{R}^d$, which takes the different characteristics of each time series into account when initialising the state, thus reducing the need to rely on a long observation history \cite{montero2021principles}. 

\paragraph{Message passing layers}
\label{sec:fibmap_mp_layers}
The \gls{mpnn} layers for each gate in the \gls{grnn}, denoted as $\text{MP}$, compute the hidden state of the $i$-th node as
\begin{equation}
    \mathbf{h}_{t,i}^{k+1} = \gamma^k \left( \mathbf{h}_{t, i}^k, \sum_{j\in \mathcal{N}(i)} {\rho^k(\mathbf{o}_i^k, \mathbf{o}_j^k, e_{ij})} \right),	    
\end{equation}
where $\mathbf{o}_i^k \in \mathbb{R}^{2d}$ represents the input of the gate at layer $k$, $\mathcal{N}(i)$ denotes the set of nodes connected to $i$, $\rho^k(\cdot)$ is a message function, $\gamma^k(\cdot)$ is an update function, and the summation serves as a permutation invariant aggregation over neighbouring nodes. Specifically, the message function $\rho^k(\cdot)$ is implemented as
\begin{subequations}
\begin{align}
    \mathbf{m}_{ij}^k &= \text{MLP}_{\text{msg}}^k([\mathbf{o}_i^k \| \mathbf{o}_j^k]),\\
    \alpha_{ij}^k &= \sigma(\mathbf{W}_{\text{msg}-\alpha}^k \mathbf{m}_{ij}^k), \\
    \mathbf{m}_{ij}^k &= \alpha_{ij}^k \mathbf{m}_{ij}^k,
\end{align}
\end{subequations}
where the messages along non-zero edges $e_{ij}$ are weighted according to an inferred scalar value $\alpha_{ij}^k$, which resides on the interval [0, 1]. The scalar value $\alpha_{ij}^k$ allows for anisotropic message passing, which assists in learning latent edge attributes such as the coupling strength between nodes. Messages are then aggregated at node $i$ using the summation, $\mathbf{m}_i^k = \sum_{j\in\mathcal{N}(i)} \mathbf{m}_{ij}^k$. Finally, the update function $\gamma^k(\cdot)$ is implemented with a residual skip connection as
\begin{subequations}
\begin{align}
    \mathbf{u}_{i}^k &= \text{MLP}_{\text{update}}^k([\mathbf{h}_i^k \| \mathbf{m}_i^k]),\\
    \mathbf{h}_{t,i}^{k+1} &= \mathbf{u}_i^k + \mathbf{W}_{\text{skip}}^k \mathbf{o}_i^k,
\end{align}
\end{subequations}
where $\mathbf{W}_{\text{skip}} \in \mathbb{R}^{d \times 2d}$ and $\mathbf{h}_i^{k+1} \in \mathbb{R}^d$. Note that the parameters of the \gls{mpnn} layers, $\text{MP}$, are defined separately for each gate and layer. This message-passing mechanism loosely resembles the gated graph convolution introduced in \cite{bresson2017residual}.

\paragraph{Iterative imputations}
\label{sec:fibmap_imputations}
To learn effective state space representations while processing the data sequentially, missing values must be accounted for. To do this, first and second-stage imputations as in \cite{cini2021filling} are performed iteratively within the spatiotemporal encoder block. The first stage imputation performs one-step-ahead prediction via a linear readout as
\begin{equation}
    \widehat{\mathbf{X}}_{t+1}^{(1)} = \mathbf{H}_t^K \mathbf{W}_{\text{stage-1}} + \mathbf{b}_{\text{stage-1}},
\end{equation}
which is used in place for the missing values. The second stage imputation acts as a type of regularisation, estimating the value at node $i$ using the previous hidden states, as well as the masks and first-stage imputations at the neighbouring nodes. The second stage imputation involves first computing an intermediate representation of each node, given by
\begin{equation}
    \mathbf{s}_{t+1, i} = \gamma_{\text{stage-2}} \left(\mathbf{h}_{t,i}^K, \sum_{j\in \mathcal{N}(i)} \rho_{\text{stage-2}}([\hat{\mathbf{x}}_{t+1,j}^{(1)} \| \mathbf{h}_{t,j}^K \| \mathbf{m}_{t+1, j}], e_{ij}) \right),
\end{equation}
where $\mathbf{s}_{t+1, i} \in \mathbb{R}^d$ and the message passing is implemented using a diffusion graph convolution \cite{atwood2016diffusion-convolutional}. Next, a linear readout layer performs the second stage imputation as
\begin{equation}
    \widehat{\mathbf{X}}_{t+1}^{(2)} = [\mathbf{S}_{t+1} \| \mathbf{H}_t^K] \mathbf{W}_{\text{stage-2}} + \mathbf{b}_{\text{stage-2}},
\end{equation}
which is used again in place of the missing values in $\mathbf{X}_{t+1}$.

The process detailed so far in the spatiotemporal encoding block is repeated for $\mathbf{X}_{t+1}$, to learn representations $\mathbf{S}_{t+2}$ and $\mathbf{H}_{t+1}^K$, and so forth, until a set of representations $\mathbf{S}_{t:t+W}$ and $\mathbf{H}_{t:t+W}^K$, and imputations $\widehat{\mathbf{X}}_{t:t+W}^{(1)}$ and $\widehat{\mathbf{X}}_{t:t+W}^{(2)}$ are calculated for the whole window length $W$. To summarise the spatiotemporal encoding block detailed in this section, we use the following shorthand notation
\begin{equation}
    \langle \mathbf{S}_{t:t+W}, \mathbf{H}_{t:t+W}^K, \widehat{\mathbf{X}}_{t:t+W}^{(1)}, \widehat{\mathbf{X}}_{t:t+W}^{(2)} \rangle = \text{ST-Encoder}(\mathbf{X}_{t:t+W}^{(p)}, \mathbf{M}_{t:t+W}^{(p)}, \mathbf{g}^{(p)}, \mathbf{V}^{(p)}).
\end{equation}

\subsubsection{The decoder}
\label{sec:fibmap_decoder}
The sequences are encoded in both the forward and backward directions, to form
\begin{equation}
    \langle \mathbf{S}_{\text{fwd}}, \mathbf{H}_{\text{fwd}}^K, \widehat{\mathbf{X}}_{\text{fwd}}^{(1)}, \widehat{\mathbf{X}}_{\text{fwd}}^{(2)} \rangle = \text{ST-Encoder}(\mathbf{X}_{t:t+W}^{(p)}, \mathbf{M}_{t:t+W}^{(p)}, \mathbf{g}^{(p)}, \mathbf{V}^{(p)}),
\end{equation}
and 
\begin{equation}
    \langle \mathbf{S}_{\text{bwd}}, \mathbf{H}_{\text{bwd}}^K, \widehat{\mathbf{X}}_{\text{bwd}}^{(1)}, \widehat{\mathbf{X}}_{\text{bwd}}^{(2)} \rangle = \text{ST-Encoder}(\mathbf{X}_{t+W:t}^{(p)}, \mathbf{M}_{t+W:t}^{(p)}, \mathbf{g}^{(p)}, \mathbf{V}^{(p)}),
\end{equation}
respectively, where $t+W:t$ denotes the time reversed/backward sequence. The next stage is to decode the hidden state representations from the encodings in both directions, to perform a final imputation at time $t^\prime$. This is done with the following function
\begin{equation}
    \widehat{\mathbf{Y}}_{t^\prime} = \text{MLP}_{\text{dec}}([\mathbf{S}_{\text{fwd}\leq t^\prime} \| \mathbf{H}_{\text{fwd}\leq t^\prime}^K \| \mathbf{S}_{\text{bwd}\geq t^\prime} \| \mathbf{H}_{\text{bwd}\geq t^\prime}^K \| \mathbf{M}_{t^\prime}^{p} \| \mathbf{V}^{(p)} \| \mathbf{g}^{(p)}]),
\end{equation}
where the notation $\text{fwd} \leq t^\prime$ refers to an index of the hidden states at time $t^\prime$ (which includes information from times $\leq t^\prime$) and similarly, $\text{bwd} \geq t^\prime$ refers to index at time $t^\prime$ (which includes information from times $\geq t^\prime$). The decoder takes a form similar to that in \cite{cini2021filling} except patient-specific parameters are also provided to facilitate personalisation. We denote the final imputed values for the entire window length as $\widehat{\mathbf{Y}}_{t:t+W}$.

\subsection{Optimisation}
\label{sec:fibmap_optimisation}
To quantify the uncertainty of the reconstructed dynamics, \textsc{FibMap} is formulated as a quantile regression \cite{koenker2001quantile}. In general, a quantile regression aims to estimate the conditional quantile function, $Q_Y(\tau|\mathbf{B})$, which represents the $\tau$-th quantile of the response variable, $Y$, given the predictor variables, $\mathbf{B}$. This is given by $Q_Y(\tau|\mathbf{B}) = \inf\{y \in \mathbb{R} | P(Y \leq y | \mathbf{B}) \geq \tau \}$, where $P(Y \leq y | \mathbf{B})$ is the conditional cumulative distribution function of the response variable, $Y$, given the predictors $\mathbf{B}$, and $\inf$ represents the infimum of the set.

In this work, conditional quantile functions for $\tau \in \mathcal{C}$, where $\mathcal{C}=\{\tau_1,\tau_2,\ldots,\tau_{|\mathcal{C}|}\}$, are predicted for each of the imputed values. This is done by computing the following masked quantile/pinball loss function
\begin{equation}
    \label{eqn:quantile_regressor_loss}
    \mathcal{L}(\widehat{\mathbf{Y}}_{t:t+T}, \widetilde{\mathbf{Y}}_{t:t+T}, \bar{\mathbf{M}}_{t:t+T}) = \frac{\sum_{h=t}^{t+T} \sum_{i=1}^N \bar{m}_{t,i} \mathcal{L}_{t,i} (\hat{\mathbf{y}}_{t,i}, y_{t,i})}{\sum_{h=t}^{t+T} \sum_{i=1}^N \bar{m}_{t,i}},
\end{equation}
where $\widehat{\mathbf{Y}}_{t:t+T} \in \mathbb{R}^{N \times |\mathcal{C}|}$, $\mathbf{Y}_t \in \mathbb{R}^{N\times 1}$, and $\bar{\mathbf{M}}_t \in \{0,1\}^{N\times 1}$, represent the predicted values, target values, and the evaluation mask, respectively, at time $t$. The function $\mathcal{L}_{t,i}$ computes the average pinball loss computed at time $t$ and node $i$, which is given by
\begin{equation}
\label{eqn:quantile_regressor_loss_internal}
\mathcal{L}_{t,i} (\hat{\mathbf{y}}_{t,i}, y_{t,i}) = \frac{1}{|\mathcal{C}|} \sum_{c=1}^{|\mathcal{C}|} (y_{t,i} - \hat{\mathbf{y}}_{t,i}[c])(\tau_c - \mathbbm{1}\{ y_{t,i} - \hat{\mathbf{y}}_{t,i}[c] < 0\}),
\end{equation}
where $c$ refers to the channel of $\hat{\mathbf{y}}_{t,i}$ (prediction vector at time $t$ and node $i$) which contains the prediction of the $c$-th conditional quantile, $Q_y( \tau_c | \ldots)$ at time $t$ and node $i$, and $\mathbbm{1}\{ y_{t,i} - \hat{\mathbf{y}}_{t,i}[c] < 0\}$ represents the indicator function, which is equal to 1 if $y_{t,i} - \hat{\mathbf{y}}_{t,i}[c] < 0$, else it is equal to $0$. In practice, we modify the decoder to predict a value for each quantile (rather than a single value) and compute the average pinball loss from each predicted quantile value as in \eqref{eqn:quantile_regressor_loss_internal}.

For \textsc{FibMap}, the following loss function is minimised
\begin{align}
\mathcal{L}_{\text{loss}} &= \mathcal{L} (\widehat{\mathbf{Y}}_{t:t+T}, \mathbf{X}_{t:t+T}, \bar{\mathbf{M}}_{t:t+T}) \\
&+ \mathcal{L} ( \widehat{\mathbf{X}}_{t:t+T}^{(1), \text{fwd}}, \mathbf{X}_{t:t+T}, \bar{\mathbf{M}}_{t:t+T} ) + \mathcal{L} ( \widehat{\mathbf{X}}_{t:t+T}^{(2), \text{fwd}}, \mathbf{X}_{t:t+T}, \bar{\mathbf{M}}_{t:t+T} ) \\
&+ \mathcal{L} ( \widehat{\mathbf{X}}_{t:t+T}^{(1), \text{bwd}}, \mathbf{X}_{t:t+T}, \bar{\mathbf{M}}_{t:t+T} ) + \mathcal{L} ( \widehat{\mathbf{X}}_{t:t+T}^{(1), \text{bwd}}, \mathbf{X}_{t:t+T}, \bar{\mathbf{M}}_{t:t+T}),
\label{eqn:combined_loss}
\end{align}
where $\mathcal{L}$ is given in \eqref{eqn:quantile_regressor_loss} and each component of the loss is specific to a different imputation stage and processing direction. The specification of the evaluation mask, $\bar{\mathbf{M}}$, differs depending on training and fine-tuning as detailed in the next sections.

\section{Data and implementation details}
\label{sec:data_and_implem}
\subsection{Dataset}
\label{sec:data_and_implem_dataset}
Our dataset comprises recordings from 51 patients with persistent \gls{af}, obtained using the AcQMap system. The system uses non-contact electrodes to sense intra-cavitary electrograms, from which dipole density measurements (charge density in Coulombs/cm) are inversely derived across the entire atria \cite{de2022critical, shi2020validation}. Each recording samples approximately 3500 nodes at 3000 Hz for 5-20 seconds. All recordings were obtained prior to pulmonary vein isolation at two United Kingdom centres between 2016 and 2023. The patient cohort (mean age 64 $\pm$ 11 years, 69\% male) had standard cardiovascular risk factors and were on guideline-directed medical therapy (see \autoref{tab:patient_characteristics} for more details).

\begin{table}[t]
    \centering
    \setlength{\tabcolsep}{25pt}
    \caption{Patient demographics and clinical characteristics. Values are mean $\pm$ SD or $n$ (\%). The asterisk denotes the presence of missing values; the number missing is shown as $n^\ast$.}
    \label{tab:patient_characteristics}
    \begin{tabular}{@{}ll@{}}
        \toprule
         & \textbf{Total ($N=51$)} \\
        \midrule
        \textbf{Demographics} & \\
        \cmidrule[0.2pt]{1-2}
        Mean age (years) & 64 $\pm$ 11 \\[2pt]
        Male sex & 35 (69\%) \\[2pt]
        Mean BMI & 29 $\pm$ 5 \\[4pt]
        \midrule
        \textbf{Comorbidities} & \\
        \cmidrule[0.2pt]{1-2}
        Hypertension & 20 (39\%) \\[2pt]
        Diabetes mellitus & 5 (10\%) \\[2pt]
        Heart failure & 17 (33\%) \\[2pt]
        CHA\textsubscript{2}DS\textsubscript{2}-VASc score $>$2 & 14 (27\%) \\[4pt]
        \midrule
        \textbf{Medications} & \\
        \cmidrule[0.2pt]{1-2}
        Beta-blockers & 33$^\ast$ (87\%, $n^\ast=13$) \\[2pt]
        Amiodarone & 11$^\ast$ (29\%, $n^\ast=13$) \\[2pt]
        Anticoagulants & 38$^\ast$ (100\%, $n^\ast=13$) \\[2pt]
        Statins & 16$^\ast$ (42\%, $n^\ast=13$) \\
        \bottomrule
    \end{tabular}
\end{table}

For pre-processing, these signals are first resampled spatially to a resolution consisting of 500 nodes. This is done by $k$-means clustering of the 3D node coordinates into $k=500$ non-overlapping clusters, where signals are resampled using the mean average signal within each cluster for each time point. Temporal resampling to 70 Hz is conducted through a combination of low pass filtering and down sampling. Low-pass filtering is applied to each time series at 70 Hz to prevent aliasing, and then the filtered signal is downsampled by reducing the sampling rate proportionally. Finally, the time series are normalised across space and time by applying min-max normalisation, where the minimum and maximum values are determined across all nodes and times, ensuring that the amplitude of the resampled signals is scaled consistently between 0 and 1.

The dataset was stratified to ensure a spectrum of \gls{af} dynamics within training (70\%), validation (10\%), and test (20\%) sets. The Shannon entropy was first computed for the resampled time series at each node to do this. The \gls{cdf} of Shannon entropy across the atrium was then created for each patient, serving as a measure to compare the similarity of the spatiotemporal dynamics between patients. The similarity between patients was assessed by computing \gls{ks} distance between \glspl{cdf} for each pair of patients, where a smaller \gls{ks} distance represents more similar entropy distributions/spatiotemporal dynamics. From this, groups of similar patients were identified by clustering a Laplacian eigenmap \cite{belkin2003laplacian} using $k$-means, where the number of clusters was determined using the elbow method. This involved forming a weighted adjacency matrix by applying a radial basis function kernel to the reciprocal of the \gls{ks} distances and thresholding. Finally, the training, validation, and test sets were formed by performing stratified sampling across these clusters, maintaining a consistent spectrum of \gls{af} dynamics within each set.

\subsection{Simulating sequential contact mapping}
\label{sec:data_and_impem_simulating}
A sequential contact mapping strategy is simulated from the non-contact recordings as a self-avoiding walk of the multipolar catheter (represented as a patch of observations) on the atrial surface; whereby the catheter surface area, dwell time, and spatial overlap, could be controlled. The self-avoiding walk is defined by first providing the catheter surface area as a fraction of the total area, where its reciprocal (1/area) is rounded up to the nearest integer to give the number of patches required to sample the whole atrium. Then, the atrium is split into disjoint patches by $k$-means clustering the 3D node coordinates, with $k$ equal to the number of patches. Spatial overlap between patches is simulated by adding additional clusters between adjacent patches and sharing node sets in proportion to the overlap required. A self-avoiding random walk ensures the path does not revisit the same region twice. This is simulated by sampling a patch randomly, then sampling the remaining patches without replacement with a probability proportional to the distance between the current and remaining patches. The resulting sequence of patches is converted to an observation mask, $\mathbf{M}$, by assigning a unity value if nodes are within the observed patch, zero otherwise, and repeating these values such that each patch is observed for a duration equal to the specified dwell time. Except during the sensitivity analysis, the sequence of patches is repeated until the length of the available recording is met.

\subsection{Training \textsc{FibMap}}
\label{sec:data_and_implem_training}
The training procedure of \textsc{FibMap} leverages the whole atria ground truth signals of each patient to learn a robust function for reconstructing whole atrium dynamics from data collected in a sequential contact mapping fashion. To do this, a self-supervised approach is employed, wherein self-avoiding walks of the multipolar catheter are sampled at random to form the observation mask, $\mathbf{M}$, and observed input data, $\mathbf{X}$, of each training sample, alongside a range of catheter surface areas (2.5 – 50\% of the atrium), dwell times (0.2 - 4 seconds) and spatial overlaps (0 – 3 additional clusters between adjacent patches). At each training iteration, batches of size $B$ are formed by collating samples from different patients, whereby $B$ inputs $\mathbf{X}_{t:t+W}^{(p)}, \mathbf{M}_{t:t+W}^{(p)}, \mathbf{A}^{(p)}, \mathbf{V}^{(p)}, \mathbf{g}^{(p)}$ are collated for a temporal input window size, $W$, sampled from the original time series following a sliding window approach with unit stride.

\begin{figure}
    \centering
    \includegraphics[width=\linewidth]{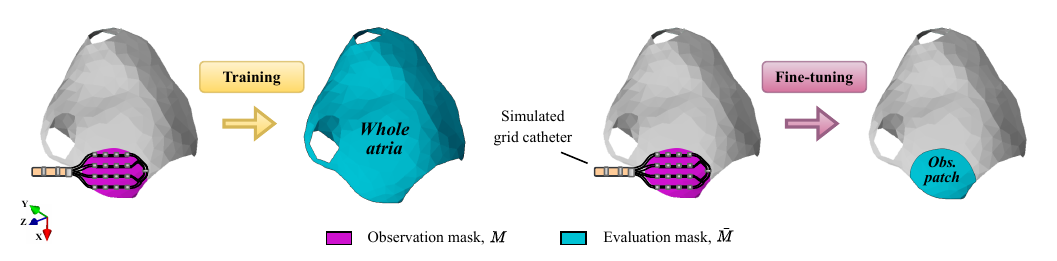}
    \caption{Definition of the observation and evaluation masks in \textsc{FibMap} training and fine-tuning. During training, \textsc{FibMap} is optimised to perform whole atria mapping using a whole atria evaluation mask to compute the loss function. During fine-tuning, only the patient-specific parameters of \textsc{FibMap} are optimised using an observed patch reconstruction loss.}
    \label{fig:fibmap_7}
\end{figure}

Imputation is performed within the temporal window and a whole atria reconstruction loss is used to optimise \textsc{FibMap} during training, whereby \eqref{eqn:combined_loss} is optimised using an evaluation mask, $\bar{\mathbf{M}}_{t:t+W}$, with all nodes and times in the input window equal to unity (see \autoref{fig:fibmap_7}). This approach ensures a robust function for reconstructing whole atria maps that generalise across the distribution of multipolar catheter paths and parameters of sequential contact mapping such as dwell time.

A hyperparameter search is conducted for the parameters shown in \autoref{tab:fibmap_hyperparameters}, by first splitting the time series of each patient sequentially, with the first 85\% of time steps being used for training, the second 5\% for validation, and the final 10\% for testing. For the validation and test sequences, a fixed self-avoiding walk of the multipolar catheter is used with a surface area of 10\%, a dwell time of 1 second, and no spatial overlap. Each configuration in the hyperparameter search is conducted for 100 epochs, whereby the \gls{mae} across the remaining atria for the validation sequence is monitored with early stopping. All experiments were performed on a NVIDIA RTX A5000 graphics processing unit. The best-performing set of hyperparameters are chosen by computing the \gls{mae} loss across the remaining atria for the test sequence and selecting the minimum loss. The best-performing configuration for training \textsc{FibMap} was found to be $B=16$, $W=40$, $d=64$, $q=64$, $r=16$, $K=1$, and 1024 hidden layer neurons. This configuration was retrained for 500 epochs, where a learning rate of $0.0009$ and a cosine scheduler were used. Training took a total of 31 hours. The result is a pre-trained FibMap model, which performs accurate and robust whole atria reconstruction from partial observations across a spectrum of dynamics found in patients and variations in multipolar mapping.

\subsection{Fine-tuning \textsc{FibMap}}
\label{sec:data_and_implem_finetuning}
In clinical practice, only the patches of multipolar catheter measurements are available from sequential contact mapping. The rest of the atrium remains unobserved. To make \textsc{FibMap} a feasible clinical solution for imputation mapping, a fine-tuning procedure is introduced to enable accurate whole atria reconstruction on new and unseen patients when only partial observations of the dynamics are available.

Our fine-tuning procedure preserves essential knowledge for whole atria reconstruction acquired during training by fixing the parameters of the pre-trained \textsc{FibMap} model, while quickly personalising the model to new patients by optimising only the node and patient embedding parameters using an observed patch reconstruction loss (see \autoref{fig:fibmap_7} for an illustration of the observation and evaluation masks used during fine-tuning, which are defined to be equal). The observed patch reconstruction loss is defined using this evaluation mask in \eqref{eqn:combined_loss}. This restricts the optimisation during fine-tuning to only the patches of multipolar catheter measurements.

Our fine-tuning procedure was configured on the validation set, which also has ground truth whole atria signals available for patients. Validation of the fine-tuning procedures aims to evaluate the relationship between the observed patch and whole atria reconstruction losses. Again, the time series of each patient was split sequentially, with the first 85\% of time steps being used for training, the second 5\% for validation, and the final 10\% for testing.

\begin{table}[t]
    \centering
    \setlength{\tabcolsep}{20pt}  
    \caption{Hyperparameter values tested during \textsc{FibMap} training.}
    \label{tab:fibmap_hyperparameters}
    \begin{tabular}{@{}ll@{}}
        \toprule
        \textbf{Hyperparameter} & \textbf{Range tested} \\
        \midrule
        Batch size, $B$ & [16, 32, 64] \\[4pt]
        Input window length, $W$ & [20, 30, 40, 50] \\[4pt]
        Hidden size, $d$ & [64, 128, 256] \\[4pt]
        Node embedding size, $q$ & [16, 32, 64] \\[4pt]
        Patient embedding size, $r$ & [16, 32, 64] \\[4pt]
        Layers, $K$ & [1, 2, 3] \\[4pt]
        Hidden layer neurons in MLP\textsubscript{enc}, MLP\textsubscript{dec} & [128, 256, 512, 1024] \\
        \bottomrule
    \end{tabular}
\end{table}

A random hyperparameter search was conducted across learning rates [0.0005, 0.005] and batch sizes [16, 32, 64, 128]. Each learning rate in the hyperparameter search was conducted for 100 epochs, whereby the \gls{mae} across the remaining atria for the validation sequence was monitored with early stopping. For the validation and test sequences, a fixed self-avoiding walk of the multipolar catheter was used with a surface area of 10\%, a dwell time of 1 second, and no spatial overlap. The best-performing set of hyperparameters was chosen by computing the \gls{mae} across the remaining atria for the test sequence and selecting the minimum loss. The best-performing configuration for fine-tuning \textsc{FibMap} was found to have a learning rate of 0.005 and a batch size of 16.

Finally, the performance of \textsc{FibMap} imputation mapping on new and unseen patients, through our fine-tuning setup, was quantified on the test set patients. A sequential time series split was not used during testing, instead, all observed measurements were used. \textsc{FibMap} was fine-tuned for 100 epochs using the configuration found in validation, and the observed patch reconstruction loss was monitored to perform early stopping. The performance metrics, detailed next, were computed across the remaining atria to assess test set performance.

\subsection{Performance metrics}
\label{sec:data_and_implem_metrics}
Quantitative assessment of the reconstructed imputation maps was performed using several metrics: \gls{mae}, \gls{mre} and \gls{mape}. These metrics evaluate the fidelity of the reconstructed whole atria maps, $\widehat{\mathbf{Y}}$, against the ground truth, $\mathbf{X}$, for the mask, $\bar{\mathbf{M}}$, which represents the logical binary complement of the observation mask, $\mathbf{M}$. The averaged performance metrics were computed via \eqref{eqn:quantile_regressor_loss}, where $\mathcal{L}_{t,i}$ was computed using each of our evaluation metrics. Metrics were computed for each model trained or fine-tuned across 5 different seeds. The average and standard deviation across these seeds were reported for each metric.

Additionally, the \gls{ps} \gls{tpr} was used to evaluate the accuracy of tracking \glspl{ps} in the reconstructed phase maps compared to the ground truth. For each patient in the test set and each imputation model, \glspl{ps} were manually labelled by two independent trained observers using a custom graphical user interface. Annotating each frame for each patient and model is labour-intensive, so only the central 70 frames (1 second) of each reconstructed phase map were annotated. Annotating \glspl{ps} for each model seed is also infeasible, so the mean and standard deviation were computed from 1000 bootstrap samples. A \glspl{ps} is considered detected if its location in the reconstructed map falls within a specified tolerance of 0.1 seconds and a 4-hop neighbourhood in space compared to its location in the ground truth map.

\subsection{Baseline models}
\label{sec:data_and_implem_baselines}
We introduce additional baseline models for the imputation mapping task: 
\begin{enumerate}
    \item Mean, which performs imputation using the node-level average;
    \item \gls{mf} with rank = 10;
    \item Univariate \gls{rnn}, which performs imputation based solely on the node-level signals;
    \item Univariate \gls{bi}-\gls{rnn}.
\end{enumerate}
Mean and \gls{mf} baseline models are employed solely on the test set due to their transductive nature. Both the univariate \gls{rnn} and \gls{bi}-\gls{rnn} models were trained using \gls{mae} loss function and followed identical hyperparameter settings and training-test protocols as \textsc{FibMap}. Unlike \textsc{FibMap}, these models did not require fine-tuning due to their inductive nature.

\subsection{Imputation mapping from EnSite Precision HD Grid recordings}
\label{sec:data_and_implem_hd_grid}
From our test cohort, three \gls{af} patients had both EnSite Precision HD Grid and AcQMap recordings collected non-contemporaneously before ablation. For these patients, AcQMap recordings were 20 seconds in duration, while EnSite Precision HD Grid recordings were significantly longer (14, 19, and 20 minutes). EnSite Precision HD Grid data consists of sequential contact mapping recordings from 16 electrodes arranged in a grid array, along with the roving 3D coordinates of each electrode within the atrium.

To ensure compatibility with \textsc{FibMap}, the EnSite Precision HD Grid recordings were pre-processed. Sparse electrogram recordings were first mapped to a uniform discretisation of the atrial surface using a nearest-neighbour approach with a 3 mm radius to interpolate the signals between electrodes. An observation mask identified active recording periods, after which signals were normalised using the maximum peak-to-peak voltage. The data was then resampled spatially to a resolution of 500 nodes using k-means clustering and resampled temporally to 70 Hz using a combination of low-pass filtering and downsampling. A final min-max normalisation ensured all signals fell within 0 and 1. From these processed measurements, imputation maps were generated using our fine-tuning procedure, whereby the observed patch reconstruction loss was monitored to perform early stopping.

We developed a sliding window cross-correlation framework to compare \textsc{FibMap} reconstructions against AcQMap `ground truth' recordings. While temporal alignment was not possible between the non-contemporaneous recordings, spatial alignment was achieved between AcQMap and imputation map by centring the vertices of the geometries around the origin, applying rigid registration using the iterative closest point algorithm \cite{besl1992method}, and projecting data between geometries using a $k=5$ nearest-neighbours regression. Using sliding window lengths from 0.5 to 4.0 seconds and a constant stride of 0.1 seconds, we computed the Pearson correlations between Hilbert phases of processed signals across all nodes and times within window pairs. This generated a cross-correlation matrix characterising the spatiotemporal similarity between recordings, which were flattened and plotted as distributions for analysis.

To validate that \textsc{FibMap} captured patient-specific dynamics, we performed three types of correlations: intra-patient comparisons between AcQMap and \textsc{FibMap} from the same patient; inter-patient comparisons between AcQMap and \textsc{FibMap} from different patients; and random baseline intra-patient comparisons between AcQMap and a spatiotemporally shuffled imputation map. For each patient, the 99th percentile of the intra, inter and shuffled distributions were plotted and statistical significance was assessed by computing the confidence intervals via bootstrapping ($n=1000$ rounds of 10000 resamples).

\end{document}